\documentclass[conference]{IEEEtran}
\IEEEoverridecommandlockouts
\usepackage{cite}
\usepackage{amsmath,amssymb,amsfonts}
\usepackage{algorithmic}
\usepackage{graphicx}
\usepackage{textcomp}
\usepackage{xcolor}
\usepackage{acronym}
\usepackage{multirow}
\def\BibTeX{{\rm B\kern-.05em{\sc i\kern-.025em b}\kern-.08em
    T\kern-.1667em\lower.7ex\hbox{E}\kern-.125emX}}

\def\Input{\mathbf{X}}
\def\rInput{\hat{\mathbf{X}}}
\def\Refs{\mathbf{R}}
\def\refs{\mathbf{r}}
\def\input{\mathbf{x}}
\def\rinput{\hat{\mathbf{x}}}
\def\fspace{\mathcal{F}}
\def\rspace{\mathbb{R}}
\def\dim{\mathcal{D}}
\def\kernelfunction{\kappa}
\def\kernelmatrix{\mathbf{K}}
\def\kernelvector{\mathbf{k}}
\def\ckernelmatrix{\mathcal{K}}
\def\ckernelvector{\mathit{k}}
\def\bkernelfunction{\tilde{\kappa}}
\def\bkernelmatrix{\tilde{\mathbf{K}}}
\def\bkernelvector{\tilde{\mathbf{k}}}
\def\cbkernelmatrix{\tilde{\mathcal{K}}}
\def\cbkernelvector{\tilde{\mathit{k}}}
\def\NoItems{N}
\def\noItems{n}
\def\noRefs{M}
\def\p{\boldsymbol{\eta}}
\def\noDim{D}
\def\one{\mathbf{1}}
\def\Identity{\mathbf{I}}

\begin{document}

\newacro{CNN}[CNN]{convolutional neural network}
\newacro{ESVDD}[ESVDD]{Ellipsoidal Support Vector Data Description}
\newacro{KDA}[KDA]{Kernel Discriminant Analysis}
\newacro{KPCA}[KPCA]{Kernel Principal Component Analysis}
\newacro{LDA}[LDA]{Linear Discriminant Analysis}
\newacro{OCSVM}[OC-SVM]{One-class Support Vector Machine}
\newacro{NPT}[NPT]{Non-linear Projection Trick}
\newacro{PCA}[PCA]{Principal Component Analysis}
\newacro{RBF}[RBF]{Radial Basis Function}
\newacro{SVDD}[SVDD]{Support Vector Data Description}
\newacro{SVM}[SVM]{Support Vector Machine}

\title{Generalized Reference Kernel\\for One-class Classification

\thanks{The work of Jenni Raitoharju was funded by Academy of Finland (project 324475).}
}


\author{\IEEEauthorblockN{Jenni Raitoharju}
\IEEEauthorblockA{\textit{Programme for Environmental Information} \\
\textit{Finnish Environment Institute}, Jyväskylä, Finland}
\and
\IEEEauthorblockN{Alexandros Iosifidis}
\IEEEauthorblockA{\textit{DIGIT, Department of Electrical and Computer Engineering} \\
\textit{Aarhus University}, Aarhus, Denmark}
}

\maketitle

\begin{abstract}
In this paper, we formulate a new generalized reference kernel hoping to improve the original base kernel using a set of reference vectors. Depending on the selected reference vectors, our formulation shows similarities to approximate kernels, random mappings, and Non-linear Projection Trick. Focusing on small-scale one-class classification, our analysis and experimental results show that the new formulation provides approaches to regularize, adjust the rank, and incorporate additional information into the kernel itself, leading to improved one-class classification accuracy.
\end{abstract}


\section{Introduction}

One-class classification aims at building a model for a class by using data from this \emph{target class} only. During inference, the model may also see \emph{outliers} not belonging to the target class and it should be able to recognize that they do not fit to the model. This kind of methods are suitable for \emph{anomaly detection}, where only samples of the normal situation are available, while any kind of anomaly should be detected \cite{pimentel2014review}. 

Traditional well-known one-class classification techniques include \ac{OCSVM} \cite{scholkopf1999support} and \ac{SVDD} \cite{tax2004support}, which are commonly applied as non-linear models exploiting the \emph{kernel trick}. Various extensions of both methods have been proposed (e.g., \cite{mygdalis2017geocc, tian2018ramp, sohrab2021multimodal}), and recently also deep neural network-based variants have been proposed \cite{ruff2018deep, chong2020simple, gautam2021graph}. Some recent works have shown that also the traditional methods used on top of deep features may be useful. In \cite{sohrab2020boosting}, both \ac{SVDD} and \ac{OCSVM} were used to improve the classification accuracy of very small classes on top of deep classification. In \cite{sohn2021learning}, \ac{OCSVM} applied on top of features obtained via self-supervised representation learning was shown to yield better results than end-to-end trained deep learning approaches. 

In this work, we focus on small-scale one-class classification tasks that do not provide enough training data for deep models. Thus, we focus on traditional one-classification methods and specifically on the kernel used with the methods. Previous works have attempted to improve the kernels for one-class classification by applying multikernel approaches that combine multiple kernels \cite{gautam2019localized, guo2021multi} or have applied approximate features to allow distributed implementation \cite{miao2018distributed}. We take a different approach and formulate a new \emph{generalized reference kernel} hoping to improve the original base kernel using a set of reference vectors. Depending on the selected reference vectors, our formulation shows similarities to approximate kernels, random mappings, and \ac{NPT}. It also reveals ways to regularize, adjust the rank, and incorporate additional information into the kernel itself.

The rest of the paper is organized as follows: Section~\ref{sec:relatedwork} first briefly introduces the one-class classification methods, \ac{SVDD} and \ac{OCSVM}, used in the experiments and then reviews works on kernel-based methods, in particular different approximate kernel methods, random features, and \ac{NPT}, which all have similarities to our proposed kernel formulation. Section~\ref{sec:proposed} introduces the proposed generalized reference mapping and kernel formulations. Section~\ref{sec:experiments} provides our experimental results and Section~\ref{sec:conclusion} concludes the paper.

\section{Related Work}
\label{sec:relatedwork}

\subsection{One-class Classification}

In this paper, we focus a one-class classification scenario with a training set $\Input = [\input_1, \dots, \input_{\noItems}] \in \rspace^{\noDim \times \NoItems}$ and a test set with  $\rInput = [\rinput_1, \dots, \rinput_{\hat{\NoItems}}] \in \rspace^{\noDim \times {\hat{\NoItems}}}$, where $\noDim$ is the data dimension and $\NoItems, {\hat{\NoItems}}$ are the number of training and testing data samples. All the samples in the training set belong to same class of interest or a \emph{target class}, and the goal is to use this data to build a model that can predict whether an unseen test sample belongs to the target class or is an \emph{outlier}.

\ac{SVDD} \cite{tax2004support} aims at enclosing the training data inside a minimum hypersphere by minimizing the following objective function: 
\begin{align}
\label{eq:SVDD}
\min \quad & F(R,\mathbf{c}) = R^2 + C\sum_{i=1}^{N} \xi_i \nonumber\\
\textrm{s.t.} \quad & \|\input_i - \mathbf{c}\|_2^2 \le R^2 + \xi _i,\nonumber\\
&\xi_i \ge 0, \:\: \forall i\in\{1,\dots,\NoItems\},
\end{align}
where $R$ is the radius of the hypersphere and $\mathbf{c}\in\mathbb{R}^{\noDim}$ is the center of the hypersphere. The hyperparameter $C>0$ and the slack variables, $\xi_i$ are used for controlling the trade-off between the volume of the hypersphere and the amount of target samples allowed outside the hypersphere. The Lagrangian dual of \eqref{eq:SVDD} (for derivation, see \cite{tax2004support}) can be given as
\begin{align}
\label{eq:lagsvdd}
\max \quad &L = \sum_{i=1}^{N} \alpha_i \mathbf{x}_i^{\intercal} \mathbf{x}_i - \sum_{i}^{N}\sum_{j}^{N} \alpha_i \alpha_j \mathbf{x}_i^T \mathbf{x}_j,\nonumber\\
\textrm{s.t.} \quad & 0 \le \alpha_i \le C.
\end{align}
Solving \eqref{eq:lagsvdd} gives a value $\alpha_i$ for each training sample $\input_i$. These values define which values are inside the hypersphere ($\alpha_i = 0$), support vectors on the boundary ($0 < \alpha_i < C$), and outliers ($\alpha_i = C$). This, in turn, allows to solve $R$ and $\mathbf{c}$ and use these to classify unseen samples.

\ac{OCSVM} \cite{scholkopf1999support} aims at separating all the training data from the origin and maximizes the distance $\rho$ from this hyperplane to the origin:
\begin{align}
\label{eq:ocsvm}
\min  \quad & \frac{1}{2} ||\mathbf{w}||^2 + \frac{1}{\nu \NoItems} \sum_{i=1}^{\NoItems} \xi_i - \rho \nonumber\\
\textrm{s.t.} \quad & \mathbf{w}^T\input_i \ge \xi_i - \rho, \:\:   \nonumber\\
 \quad & \xi_i \ge 0, \:\: \forall i\in\{1,\dots,\NoItems\},
\end{align}
where $\mathbf{w}$ is a weight vector, slack variables $\xi_i$ allow some data points to lie within the margin, and hyper-parameter $\nu$ sets an upper bound on the fraction of training samples allowed within the margin. The dual of \eqref{eq:ocsvm} can be given as:
\begin{equation}
\label{eq:lagocsvm}
\min L = \frac{1}{2}\sum_{i=1}^{\NoItems} \alpha_i\alpha_i \mathbf{x}_i^T \mathbf{x}_i,
\textrm{s.t. } 0 \le \alpha_i \le \frac{1}{\nu\NoItems}, \sum_{i=1}^{\NoItems} \alpha_i = 1.
\end{equation}
Now samples on the hyperplane have $0 < \alpha_i < \frac{1}{\nu\NoItems}$ and they can be used to solve $\rho.$ 

\subsection{Kernel-based Methods}
\label{ssec:kernelmethods}

Non-linear versions of various pattern recognition techniques can be obtained by using the \emph{kernel trick}. The kernel trick was introduced already in 1960s for kernel perceptrons \cite{aizerman1964theoretical} and became popular in 1990s along with \acp{SVM} \cite{cortes1995support}, and has been combined also with state-of-the-art \acp{CNN} \cite{mairal2016end, chen2020convolutional} with promising results. 

The main idea of kernel methods is to non-linearly map the input data to a feature space employing a \emph{mapping function} $\phi(\cdot) : \input_i \in \rspace^\dim \rightarrow \phi(\input_i) \in \fspace$   
and then apply the original linear method in that space. The feature space $\fspace$ usually has the properties of Hilbert spaces \cite{scholkopf1999input,argyriou2009when} and it is often selected to be higher or even infinite-dimensional space leading to the classes to be more likely linearly separable. While this makes it infeasible to operate directly on the mapped samples, many pattern recognition techniques use only inner products of the input samples, not the samples by themselves (as examples, see \ac{SVDD} \eqref{eq:lagsvdd} and \ac{OCSVM} \eqref{eq:lagocsvm}). Thus, if the inner products are known, the mapped samples $\phi(\input_i)$ and even the function $\phi(\cdot)$ are not needed. This is the basis of the kernel trick: the inner products of the mapped samples are obtained using a \emph{kernel function}:
\begin{equation}
\kernelfunction(\input_i,\input_j) \triangleq \phi(\mathbf{x}_i)^T \phi(\mathbf{x}_j) 
\end{equation}
without explicitly using the function $\phi(\cdot)$. An example of the kernel function is the widely used \ac{RBF}: 
\begin{equation}
\label{eq:rbf}
\kernelfunction(\input_i,\input_j) =  \exp  \left( \frac{ -\| \input_i - \input_j\|_2^2 }{ 2\sigma^2 } \right),
\end{equation}
where $\sigma$ is a hyperparameter. The kernel trick has been used to create many widely-used non-linear variants of originally linear pattern recognition techniques, such \ac{KPCA} \cite{scholkopf1998kpca} from \ac{PCA}, \ac{KDA} \cite{baudat2000generalized, mika1999fisher} from \ac{LDA}, and non-linear \ac{SVDD} \cite{tax2004support}.

An inherit drawback of kernel methods for large datasets is the fact that they require the calculation of the so-called \emph{kernel matrix} $\kernelmatrix_{\Input \Input} \in \rspace^{\NoItems \times \NoItems}$, having elements $[\kernelmatrix_{\Input \Input}]_{i,j} = \kernelfunction(\input_i,\input_j)$, and the exploitation of $\kernelmatrix_{\Input \Input}$ in order to learn the model parameters. For example, \ac{KPCA} and \ac{KDA} solve a (generalized) eigendecomposition problem defined on $\kernelmatrix_{\Input \Input}$ and, thus, their time complexity is a function of $\NoItems^3$. For large scale problems, such a time complexity is prohibitive and, thus, approximate kernel approaches have been proposed to overcome this issue \cite{mahoney2011randomized, choromanski2022hybrid, gittens2016revisiting}. They aim at constructing a low-rank approximation of the kernel matrix $\kernelmatrix_{\Input \Input}$ to reduce the computational complexity of applying the main algorithm.

Some of well-known techniques to generate approximate kernels include \emph{random sampling} methods (e.g., \cite{drineas2005nystrom, drineas2006fastII, alaoui2015fast, gittens2016revisiting}), \emph{random projection} methods (e.g., \cite{ailon2009fast}), and \emph{random Fourier features} (e.g., \cite{rahimi2007random, avron2017random, liao2020random}). Random sampling or \emph{Nyström methods} sample a subset of the columns of $\kernelmatrix_{\Input \Input}$ to construct the approximate matrix. An important aspect in this process is the distribution used to sample the columns. While it would be possible to simply use a uniform distribution (vanilla Nyström), other distributions have been shown to be more efficient \cite{drineas2005nystrom}. Random projection methods multiply the original kernel matrix with a random data-independent projection matrix, thus selecting linear combinations of the columns of $\kernelmatrix_{\Input \Input}$. Random Fourier features take a different approach. Unlike random sampling and random projection methods that try to approximate the kernel matrix $\kernelmatrix_{\Input \Input}$, they try to approximate the kernel function $\kernelfunction(\input_i, \input_j)$ using a data-independent approach. 
It was demonstrated in \cite{yang2012comparison, gittens2016revisiting} that the data-dependent Nyström methods can lead to better kernel approximations than data-independent random projection methods or random Fourier features.

To give a more mathematical explanation, random sampling methods approximate the full kernel matrix $\kernelmatrix_{\Input \Input}$ by randomly selecting a subset of $\noItems$ samples, $\rInput = [\rinput_1,\dots, \rinput_m] \in \rspace^{\noDim \times \noItems}$, among the original $\NoItems$ samples and then constructing the low-rank matrix as 
\begin{equation}
\label{eq:sampling}
\hat{\kernelmatrix}_{\Input \Input} = \kernelmatrix_{\Input \rInput}  \kernelmatrix_{\rInput \rInput}^{+}  \kernelmatrix_{\rInput \Input},
\end{equation}
where $\kernelmatrix_{\Input \rInput} = [\kernelfunction(\input_i,\rinput_j)]_{ij} \in \rspace^{\NoItems \times \noItems}$, $\kernelmatrix_{\rInput \rInput} = [\kernelfunction(\rinput_i,\rinput_j)]_{ij}$ $\in \rspace^{\noItems \times \noItems}$, $\kernelmatrix_{\Input \rInput} = [\kernelfunction(\input_i,\rinput_j)]_{ij} \in \rspace^{\noItems \times \NoItems}$, and $(\cdot)^+$ denotes a pseudo inverse. Besides random sampling, the vector in $\rInput$ can be selected, e.g., via clustering the input vectors and using the cluster centroids \cite{zhang2010clustered}.

As derived in \cite{rahimi2007random, avron2017random}, random Fourier features are based on the knowledge that for shift invariate kernel functions, i.e., kernel functions of form $\kernelfunction(\input_i, \input_j) = \kernelfunction(\input_i - \input_j)$ having $\kernelfunction(0) = 1$, there is a probability distribution $p_\kernelfunction(\cdot)$ such that
\begin{equation}
\kernelfunction(\input_i, \input_j) = \int_{\rspace^D} e^{-2\pi i \p^T(\input_i - \input_j)}p_\kernelfunction(\p)d\p,
\end{equation}
i.e., the inverse Fourier transform of a kernel function $\kernelfunction(\cdot, \cdot)$ is a probability distribution $p_\kernelfunction(\cdot)$. Drawing $\p_1, \dots, \p_M$ from $p_\kernelfunction(\cdot)$ and defining
\begin{equation}
\label{eq:fourierfeature}
\phi_{\p}(\input_i) = \frac{1}{\sqrt{m}} [e^{-2\pi i \p_1^T\input_i}, \dots, e^{-2\pi i \p_m^T\input_i}]^*,
\end{equation}
leads to
\begin{equation}
\kernelfunction(\input_i, \input_j) = \mathbb{E}_{\p} [\phi_{\p}(\input_i)^T\phi_{\p}(\input_j)].
\end{equation}
For the \ac{RBF} kernel in \eqref{eq:rbf}, the corresponding probability distribution $p_\kernelfunction(\cdot)$ is $\mathcal{N}(0, \sigma^2 \mathbf{I})$ \cite{yang2012comparison} and, thus, random Fourier features can be generated by sampling vectors $\p_1, \dots, \p_M$ from this distribution and computing $\phi_{\p}(\input_i)$ according to \eqref{eq:fourierfeature}. The mapped representations $\phi_{\p}(\input_i)$ can be then used as new features for $\input_i$, which can be used with the original linear methods, or they can be used to construct a low-rank approximation (assuming $M < \NoItems$), of the kernel matrix as $\hat{\kernelmatrix}_{\Input \Input} = [\phi_{\p}(\input_i)^T\phi_{\p}(\input_j)]_{ij}$.

The optimal low-rank approximations in terms spectral or Frobenius norms can be obtained using eigenvalue decomposition \cite{li2015large}, but it has prohibitive cubic complexity. Furthermore, it is not clear what is the best way to measure the optimality of the transformation. Therefore, most of the research on approximate kernels focuses on the following aspects: 1) How to evaluate the optimality of the approximation? \cite{yang2012comparison, liao2020random, gittens2016revisiting, li2019towards}, 2) How to construct more efficiently more optimal approximations? \cite{musco2017recursive, li2015large, avron2017faster}. Some works focus on analyzing the performance of the approximate kernels in a specific application \cite{bach2013sharp, alaoui2015fast}, and some works study the rank of the low-rank approximations \cite{bach2013sharp, wang2018numerical}. Interestingly, all this research assumes that the original kernel is optimal or contains all the relevant information that can be extracted. However, this assumption has not been justified.

While the kernel trick can be used in numerous applications, there are also methods that cannot be presented in terms of inner products only (e.g., \ac{ESVDD}). In such cases, the mapping function $\phi(\cdot)$ would be directly needed to first map the input vectors to the feature space and then apply the method directly on these new features. However, for many kernel functions the corresponding feature space can be infinite dimensional and/or the mapping function can be unknown. The problem can be approached by mapping the inputs $\input_i$ to their \emph{pre-images} $\boldsymbol{\phi}_i$ \cite{kwok2004preimage, mika1998kernel}, which satisfy $\boldsymbol{\phi}_i^T\boldsymbol{\phi}_j \approx \kernelfunction(\input_i, \input_j)$. Random Fourier features \eqref{eq:fourierfeature} can be directly used are pre-images as $\boldsymbol{\phi}_i = \hat{\phi}(\input_i)$. When the computational complexity of obtaining the pre-images is not a concern, \ac{NPT} \cite{kwak2013NPT} provides an alternative solution, while its approximate version can be used for large-scale problems \cite{iosifidis2016NystromNPT}.

\ac{NPT} maps the data to the \emph{effective subspace} $\mathcal{P}$ of the feature space $\mathcal{F}$ by 
as follows: First, the original kernel matrix $\kernelmatrix_{\Input \Input} \in \rspace^{\NoItems \times \NoItems}$ is computed and centered (see Appendix \eqref{eq:center_ref}). Then, the eigendecomposition of the centered kernel matrix $\ckernelmatrix$ is computed as
\begin{equation}
\ckernelmatrix_{\Input \Input} = \mathbf{U} \boldsymbol{\Lambda} \mathbf{U}^{-1} = \mathbf{U} \boldsymbol{\Lambda} \mathbf{U}^T = \mathbf{U}_r \boldsymbol{\Lambda}_r \mathbf{U}_r^T,
\end{equation}
where $\mathbf{U}$ contains as its columns the eigenvectors, $\boldsymbol{\Lambda}$ is a diagonal matrix with the eigenvalues on its diagonal, the second equality follows from the symmetry, and in the last step only the non-zero eigenvalues and the corresponding eigenvalues are kept, i.e., $\mathbf{U}_r \in \rspace^{\NoItems \times r}$ and $\boldsymbol{\Lambda}_r \in \rspace^{r \times r}$, where $r$ is the rank of $\mathcal{K}_{\Input \Input}$. Finally, the matrix containing the pre-images, $\boldsymbol{\Phi} = [\boldsymbol{\phi}_1, \dots, \boldsymbol{\phi}_\NoItems] \in \rspace^{r \times \NoItems}$, is computed as
\begin{equation}
\label{eq:NPTtrain}
\mathbf{\Phi} = \boldsymbol{\Lambda}_r^{\frac{1}{2}} \mathbf{U}_r^T.
\end{equation}
In the test phase, the pre-image or a test vector $\hat{\input}_j$ is computed as follows: First the kernel vector $\kernelvector_{\Input \hat{\input}_j} = [\kernelfunction( \input_i, \hat{\input}_j)]_{i} \in \rspace^{\NoItems \times 1}$ is computed and centered (see Appendix \eqref{eq:center_other}). 
Finally, the pre-image of the test vector is computed as
\begin{equation}
\label{eq:NPTtest}
\hat{\boldsymbol{\phi}}_j = \boldsymbol{\Lambda}_r^{-\frac{1}{2}} \mathbf{U}_r^T \ckernelvector_{\Input \hat{\input}_j},
\end{equation}
where $\ckernelvector_{\Input \hat{\input}_j}$ is the centered kernel vector. As computing the pre-images via \ac{NPT} requires the eigendecomposition of $\ckernelmatrix_{\Input \Input}$, it suffers from the inherent problem of standard kernel methods related to their computational cost. Therefore, it is most suitable for smaller problem that cannot be solved using the kernel trick. An approximate version of \ac{NPT} \cite{iosifidis2016NystromNPT} uses a Nystr\"{o}m approximation of the full kernel matrix, making the computational cost of the eigendecomposition of the kernel matrix lower. Since both \ac{NPT} and its approximate version are based on the kernel matrix, they still rely on a similar assumption of the original kernel matrix being the optimal solution.

\section{Generalized Reference Mapping and Kernel}
\label{sec:proposed}
As discussed above, a lot of research from different perspectives has been done to construct better kernel approximations in large-scale problems. While it is known that approximate methods can leads to implicit regularization of the methods for noisy data \cite{mahoney2011implementing}, most of the above discussed research shares the same underlying assumption that the original kernel function or kernel matrix is optimal and worth approximating as closely as possible. While some studies focus on kernel function selection \cite{liu2014research, ong2005hyperkernels, motai2014kernel}, and \emph{hyperkernels} or \emph{multikernels} have been proposed to combine different kernel functions to create better kernels \cite{ong2005hyperkernels, guo2021multi}, the optimality of the kernel functions has received surprisingly little attention in general. In this paper, we take a different approach from the works discussed in Section \ref{ssec:kernelmethods} and, instead of attempting to approximate the original kernel, we formulate a new kernel definition using a set of selected reference vectors. We focus on small-scale one-class classification problems, where the computational complexity is not a concern but rather the lack of training samples. We hope to use our new kernel as an implicit data augmentation for such problems. 

In our kernel definition, we use the following notations: We consider a training set $\Input = [\input_1, \dots, \input_{\noItems}] \in \rspace^{\noDim \times \NoItems}$ and a test set $\rInput = [\rinput_1, \dots, \rinput_{\hat{\NoItems}}] \in \rspace^{\noDim \times {\hat{\NoItems}}}$. As we focus on one-class classification, the training set is assumed to contain samples from a single class only. Both training and test sets are assumed \emph{standardized} using the training set mean and standard deviation. Furthermore, our kernel uses a set of reference vectors $\Refs = [\refs_1, \dots, \refs_{\noRefs}] \in \rspace^{\noDim \times \noRefs}$, where $\noRefs$ can be less, equal, or greater than $\NoItems$. We will later return the selection of $\noRefs$ and the reference vectors. Our kernel definition is built on an original kernel to which we refer to as the \emph{base kernel}. We use a tilde to denote all the related terms (kernel matrix $\bkernelmatrix$, kernel vector $\bkernelvector$, and kernel function $\bkernelfunction(\cdot,\cdot)$), whereas the terms referring to proposed reference kernel do not have tildes. Furthermore, we use subscripts with kernel matrices and vectors to denote the data over which they have been computed as follows:
\begin{equation}
\kernelmatrix_{\Input \rInput} = [\kernelfunction(\input_i,\rinput_j)]_{ij} \in \rspace^{\NoItems \times {\hat{\NoItems}}},
\end{equation}
where $\input_i \in \Input$ and $\rinput_j \in \rInput$, and
\begin{equation}
\kernelvector_{\Input \rinput_j} = [\kernelfunction(\input_i,\rinput_j)]_{i} \in \rspace^{\NoItems \times 1},
\end{equation}
where $\input_i \in \Input$.

To formulate a new kernel using a base kernel function $\bkernelfunction(\cdot, \cdot)$ and a set of reference vectors $\Refs$, we first compute the reference base kernel matrix $\bkernelmatrix_{\Refs \Refs}$ and center it (see Appendix \eqref{eq:center_ref}). Then, we calculate the eigendecomposition of the centered base kernel matrix as $\cbkernelmatrix_{\Refs \Refs}$
\begin{equation}
\label{eq:refeigen}
\cbkernelmatrix_{\Refs \Refs} = \mathbf{U} \boldsymbol{\Lambda} \mathbf{U}^{-1} = \mathbf{U} \boldsymbol{\Lambda} \mathbf{U}^T = \mathbf{U}_r \boldsymbol{\Lambda}_r \mathbf{U}_r^T,
\end{equation}
where $\mathbf{U}$ contains as its columns the eigenvectors and $\boldsymbol{\Lambda}$ is a diagonal matrix with the eigenvalues on its diagonal. We keep only the $r$ non-zero eigenvalues $\lambda_t$ and the corresponding eigenvectors $\mathbf{u}_t$, where $r$ is the rank of $\cbkernelmatrix_{\Refs \Refs}$.

We now define our generalized reference mapping function $\phi(\cdot)$:
\begin{align}
\label{eq:refmapping}
[\phi(\input_i)]_t &= \sum_{m=1}^\noRefs \frac{1}{\lambda_t} u_{tm}  \bkernelfunction(\input_i, \refs_m) = \frac{1}{\lambda_t} \mathbf{u}_t^T  \cbkernelvector_{\Refs \input_i} \nonumber \\
\rightarrow \phi(\input_i) &=  \boldsymbol{\Lambda}_r^{-\frac{1}{2}} \mathbf{U}_r^T  \cbkernelvector_{\Refs \input_i},
\end{align}
where $\phi(\input_i) \in \rspace^r$ and $\cbkernelvector_{\Refs \input_i}$ is the centered version (using Appendix \eqref{eq:center_other}) of the base kernel vector $\bkernelvector_{\Refs \input_i}$. Using the defined mapping function $\phi(\cdot)$, we can either map all the samples into the corresponding feature space and apply the linear methods on the new features or compute the kernel matrices using the corresponding kernel function:
\begin{eqnarray}
\label{eq:refkernel}
\kernelfunction(\input_i, \input_j) &=& \phi(\input_i)^T\phi(\input_j) = \cbkernelvector_{\Refs \input_i}^T \mathbf{U}_r \boldsymbol{\Lambda}_r^{-1} \mathbf{U}_r^T  \cbkernelvector_{\Refs \input_i} \nonumber \\
&=& \cbkernelvector_{\Refs \input_i}^T \cbkernelmatrix_{\Refs \Refs}^{+}  \cbkernelvector_{\Refs \input_i},
\end{eqnarray}
where $\cbkernelmatrix_{\Refs \Refs}^{+} = \mathbf{U} \Lambda^{-1} \mathbf{U}^T$ is the pseudo inverse of $\cbkernelmatrix_{\Refs \Refs}$. Using this kernel function, the kernel matrix $\kernelmatrix_{\Input \Input}$ becomes
\begin{equation}
\label{eq:refkernelmatrix}
\kernelmatrix_{\Input \Input} = \phi(\Input)^T\phi(\Input) = \cbkernelmatrix_{\Refs \Input}^T\cbkernelmatrix_{\Refs \Refs}^{+} \cbkernelmatrix_{\Refs \Input} = \cbkernelmatrix_{\Input \Refs}\cbkernelmatrix_{\Refs \Refs}^{+} \cbkernelmatrix_{\Refs \Input}.
\end{equation}
Similarly, the test kernel matrix $\kernelmatrix_{\Input \hat{\Input}}$ becomes
\begin{equation}
\kernelmatrix_{\Input \hat{\Input}} =  \cbkernelmatrix_{\Input \Refs}\cbkernelmatrix_{\Refs \Refs}^{+} \cbkernelmatrix_{\Refs \hat{\Input}}.  
\end{equation}

As can be seen, the above definitions for reference mapping and reference kernel function are not following the typical approach, where only the kernel function is known, while $\phi(\cdot)$ is possibly both undefined and infinite-dimensional. Instead, we now have an explicit formula for $\phi(\cdot) \in \rspace^r$, where $r$ is the rank of $\cbkernelmatrix_{\Refs \Refs}$, and the formula depends on the selected reference vectors. Here, it should be noted we see the reference vectors as hyperparameters of the mapping/kernel function. They can be either data-dependent or data-independent as discussed below, but both cases are common also for the common kernel function. For example, the value of $\sigma$ in the \ac{RBF} kernel function is often scaled according to the training data. Despite the unusual definition, kernel function in \eqref{eq:refkernel} fulfils the definition of a proper kernel function \cite{bishop2006pattern}: A
necessary and sufficient condition for a function $\kernelfunction(\cdot, \cdot)$ to be a valid kernel is that the kernel matrix $\kernelmatrix_{\Input \Input}$ is
positive semidefinite. Our kernel matrix is Gram matrix by definition \eqref{eq:refkernelmatrix} and, thus, it is always positive semidefinite. 

We call our new definitions \emph{generalized reference mapping and kernel}, because different selection of the reference vectors $\Refs$ produces kernels, which are equivalent to different approximate solutions discussed in Section~\ref{ssec:kernelmethods}. In particular, we consider the following seven cases:
\begin{enumerate}
    \item $\Refs=\Input \in \rspace^{\noDim \times \NoItems}$: Here, the target class training samples are used as reference vectors, and the mapping function $\phi(\cdot)$ in \eqref{eq:refmapping} is equivalent to \ac{NPT} in \eqref{eq:NPTtrain} and \eqref{eq:NPTtest}. The kernel function $\kernelfunction(\cdot,\cdot)$ in \eqref{eq:refkernel} is almost equal to the base kernel function with some difference coming from the centering operations.    
    \item $\Refs = \mathbf{N} \in \rspace^{\noDim \times \NoItems} \sim \mathcal{N}(0,1)$: Here, $\NoItems$ random reference vectors are drawn from the normal distribution. As the training data is standardized and contains samples from the target class only, the reference vectors come approximately from the training distribution. This may have a regularizing effect. With this reference vector selection, the generalized reference mapping has connections with the mappings used by randomized single hidden-layer feedforward network one-class classifiers, e.g., \cite{leng2015one,iosifidis2017one},  with two differences, i.e., the centering of the base kernel vector $\cbkernelvector_{\Refs \input_i}$ and the normalization of the mapping $\phi(\input_i)$, which follows from the projection and normalization with the eigenpairs of $\cbkernelmatrix_{\Refs \Refs}$ in \eqref{eq:refmapping}. 
    \item $\Refs=\Input^* \in \rspace^{\noDim \times \noRefs}, \: \noRefs < \NoItems$: Here, we pick randomly a subset of the training samples. This option corresponds to the vanilla Nyström approximate kernel approach \eqref{eq:sampling}. Using this reference vector selection along with the mapping function in \eqref{eq:refmapping} can be seen as an approximate variant of \ac{NPT}.
    \item $\Refs = \mathbf{N} \in \rspace^{\noDim \times \noRefs} \sim \mathcal{N}(0,1), \: \noRefs < \NoItems$: Here, $\noRefs < \NoItems$ random reference vectors drawn from the normal distribution are used. This approach is similar to the random projection approximate kernel approaches discussed in Section~\ref{ssec:kernelmethods}. Again, using this version with the mapping function in \eqref{eq:refmapping} allows to extend the approximate kernel method to the \ac{NPT} setting.  
    \item $\Refs = [\Input, \Input_{neg}] \in \rspace^{\noDim \times (\NoItems + T)}$: Here, the reference vectors are the target class training samples augmented with some of the non-target class training samples. Note that this variant violates the basic assumption of one-class classification that no negative samples are available during training. In this paper, this case is included only to see if such data augmentation (if possible) would help. In future work, this approach of exploiting negative samples in SVDD can be compared with another conceptually similar but methodologically different approach, `\ac{SVDD} with negative examples', proposed in \cite{tax2004support} or with class-specific methods (e.g., \cite{iosifidis2016class, raitoharju2019null}).
    \item $\Refs = [\Input, \mathbf{N}] \in \rspace^{\noDim \times (\NoItems+T)}$: Here, the reference vectors are the target class training vectors augmented by $T$ random vectors drawn from normal distribution. This approach can be seen as an implicit data augmentation scheme for non-linear one-class classification.   
    \item $\Refs = \mathbf{N} \in \rspace^{\noDim \times (\NoItems+T)} \sim \mathcal{N}(0,1)$: Here, we draw $\NoItems+T$ random reference vectors from the normal distribution for direct comparison against the previous two cases.  
\end{enumerate}

\section{Experiments}
\label{sec:experiments}

\subsection{Experimental Setup}
\label{ssec:setup}

We experimentally compared the seven cases of our proposed generalized reference approach, both by using the proposed kernel function in \eqref{eq:refkernel} to create the corresponding kernel matrix to be used in the kernel implementation of the one-class classification techniques and by using the proposed mapping function in \eqref{eq:refmapping} to generate new features to be used in the linear implementation of the one-class classification techniques. The kernel approach was compared against the base kernel and the mapping approach was compared against \ac{NPT} using the base kernel function. The base kernel function in all experiments was the \ac{RBF} kernel \eqref{eq:rbf} and the one-class classification techniques used in the experiments were \ac{SVDD} and \ac{OCSVM}.  The codes for linear \ac{SVDD} and both linear and kernel \ac{OCSVM} were obtained from LIBSVM library\footnote{https://www.csie.ntu.edu.tw/~cjlin/libsvm/}. The code for kernel \ac{SVDD} was obtained from dd-tools\footnote{https://www.tudelft.nl/ewi/over-de-faculteit/afdelingen/intelligent-systems/pattern-recognition-bioinformatics/pattern-recognition-bioinformatics/data-and-software/dd-tools}. Our own codes were implemented on Matlab R2017b and are publicly available\footnote{https://github.com/JenniRaitoharju/GeneralizedReferenceKernel}.

We selected for our experiments six different small-scale datasets from UCI machine learning repository \cite{dua2019uci} shown in Table \ref{tab:datasets}. For each class in these datasets, we created a different one-class classification task by considering this class as the target class and all the other classes as outliers.  Thus, we had a total of 14 different tasks as shown in Table \ref{tab:datasets}. For each task, we selected randomly 70\% of the samples as our training set and the remaining 30\% as the test set. This was repeated five times and the same splits were used to test the performance of all methods. For each split, the experiments were also repeated five times. Thus, each reported result is an average of 25 different runs. For each run, the hyperparameters were selected using a random 5-fold cross-validation approach within the training set. Thus, also for non-random kernel/mappings variants, there may be differences between the repetitions due to the differences in validation splits that may lead to different hyperparameter selection.

\begin{table}[tbp]
\setlength{\tabcolsep}{4pt}
\caption{Dataset properties}
\vspace{-5pt}
\label{tab:datasets}
\begin{center}
\begin{tabular}{|c|ccccccc|}
\hline
\bf{Dataset}& $C$ & $N_{tot}$ & $D$ & Task & Target & $N$ & $T$\\
\hline
\multirow{3}{*}{Iris} & \multirow{3}{*}{3} & \multirow{3}{*}{150} & \multirow{3}{*}{4} & Iris1 & Setosa & 35 & 35\\
&&&& Iris2 & Versicolor & 35 & 35\\
&&&& Iris3 & Virginica & 35 & 35\\
\hline
\multirow{3}{*}{Seeds} & \multirow{3}{*}{3} & \multirow{3}{*}{210} & \multirow{3}{*}{7} & Seed1 & Kama & 49 & 49\\
&&&& Seed2 & Rosa & 49 & 49\\
&&&& Seed3 & Canadian & 49 & 49\\
\hline
\multirow{2}{*}{Ionosphere} & \multirow{2}{*}{2} & \multirow{2}{*}{351} & \multirow{2}{*}{34} & Ion1 & Bad & 88/89 & 88/89\\
&&&& Ion2 & Good & 157/158 & 88/89\\
\hline
\multirow{2}{*}{Sonar} & \multirow{2}{*}{2} & \multirow{2}{*}{208} & \multirow{2}{*}{60} & Son1 & Rock & 68 & 68\\
&&&& Son2 & Mines & 78 & 68\\
\hline
Qualitative  & \multirow{2}{*}{2} & \multirow{2}{*}{250} & \multirow{2}{*}{6} & Bank1 & Bankr. & 74/75 & 74/75\\
bankruptcy &&&& Bank2 & No bankr. & 100/101 & 74/75\\
\hline
Somerville  & \multirow{2}{*}{2} & \multirow{2}{*}{143} & \multirow{2}{*}{6} & Happ1 & Happy & 54 & 47\\
happiness &&&& Happ2 & Unhappy & 47 & 47\\
\hline
\multicolumn{8}{l}{$C$ - number of classes, $N_{tot}$- total number of samples, }\\
\multicolumn{8}{l}{
$D$ - dimensionality, Task - subtask abbreviation, Target - target class}\\
\multicolumn{8}{l}{
in the subtask, $N$, $T$ - as defined in Section \ref{sec:proposed}}
\end{tabular}
\end{center}
\end{table}

We experimented with the cases 3-4 listed in Section \ref{sec:proposed} by setting $\noRefs = \lfloor \frac{\NoItems}{2}\rfloor$ and with cases 5-7 by setting $T = \min(\NoItems, \NoItems_{neg})$, i.e., in these cases we doubled the number of reference vectors, if the number of negative samples allowed it. While the number of negative samples is not relevant for cases 6-7, we used the same $T$ to allow direct comparison. The number of training samples, $\NoItems$, and the value of $T$ used for cases 5-7 are also shown in Table~\ref{tab:datasets}. When forming $\mathbf{U}_r$ and $\boldsymbol{\Lambda}_r$ in \eqref{eq:refeigen} we consider values $<10^{-6}$ as zeros.
The hyperparameter values selected by cross-validation included $\sigma$ for \ac{RBF} kernel, $C$ for \ac{SVDD}, and $\nu$ for \ac{OCSVM}. Hyperparameter $\sigma$ was set as $\sqrt{sd_{aver}}$, were $d_{aver}$ is the average squared distance been the training samples and $s$ is selected from the following options: $[10^{-1}, 10^0, 10^1, 10^2, 10^3]$. Hyperparameters $C$ and $\nu$ are selected from $[0.1, 0.2, 0.3, 0.4, 0.5, 0.6]$. As the performance metric, we use Geometric Mean (Gmean), because it takes into account both True Positive Rate (TPR) and True Negative Rate (TNR) as $ \text{Gmean}=\sqrt {\text{TPR} \times \text{TNR}}.$

\subsection{Experimental Results and Discussion}

\setlength{\tabcolsep}{2pt}

\begin{table*}[htbp]
\caption{Average Gmean values (and kernel matrix ranks) for \ac{SVDD} with the base kernel function and the seven cases of our proposed generalized reference kernel function}
\label{tab:SVDDkernel}
\begin{center}
\resizebox{1\linewidth}{!}{
\begin{tabular}{|c||c|c|c||c|c||c|c|c|}
\hline
&\multicolumn{3}{|c||}{\textbf{$M=N$}}&\multicolumn{2}{c||}{\textbf{$M=\lfloor N/2 \rfloor$}}&\multicolumn{3}{c|}{\textbf{$M=N+T$}} \\
\cline{2-9} 
\textbf{Data} & \textbf{Base} & \textbf{Case 1} & \textbf{Case 2} & \textbf{Case 3} & \textbf{Case 4} & \textbf{Case 5} & \textbf{Case 6} & \textbf{Case 7}  \\
\hline
Iris1 & 80.3$\pm$9.6 (34.4) & 80.3$\pm$9.6 (18.4) & \textbf{90.2$\pm$4.3} (23.7) & 86.2$\pm$8.0 (12.4) & \textbf{89.9$\pm$4.6} (14.8) & \textbf{90.0$\pm$3.1} (34.0) & 88.4$\pm$5.6 (16.5) & 86.2$\pm$8.1 (19.0) \\ 
Iris2 & \textbf{91.0$\pm$4.9} (35.0) & \textbf{91.0$\pm$4.9} (21.8) & 90.9$\pm$5.0 (24.7) & 90.0$\pm$5.4 (13.1) & \textbf{91.0$\pm$4.9} (15.8) & 89.5$\pm$4.6 (35.0) & \textbf{91.2$\pm$4.9} (32.1) & 90.3$\pm$5.7 (31.1) \\ 
Iris3& 88.0$\pm$4.5 (34.8) & 88.6$\pm$4.7 (23.3) & \textbf{90.3$\pm$3.0} (27.3) & \textbf{89.0$\pm$3.8} (14.2) & \textbf{89.0$\pm$4.0} (14.9) & \textbf{89.0$\pm$4.8} (34.8) & 86.8$\pm$8.7 (23.8) & 86.5$\pm$9.4 (28.0) \\ 
\hline
Seed1 & 86.1$\pm$3.7 (47.1) & 87.2$\pm$3.2 (21.6) & \textbf{87.4$\pm$2.9} (28.6) & 86.5$\pm$4.4 (15.6) & \textbf{87.2$\pm$3.4} (19.2) & \textbf{88.7$\pm$3.3} (49.0) & 87.2$\pm$2.4 (30.7) & 86.9$\pm$4.2 (36.2) \\ 
Seed2 & \textbf{92.3$\pm$8.5} (48.8) & \textbf{92.3$\pm$8.5} (31.4) & 91.6$\pm$8.4 (40.1) & \textbf{93.6$\pm$2.9} (19.2) & 92.2$\pm$8.5 (22.4) & 90.5$\pm$2.7 (49.0) & 91.9$\pm$8.4 (47.4) & \textbf{92.2$\pm$8.5} (40.3) \\ 
Seed3 & 87.0$\pm$10.6 (49.0) & 87.1$\pm$10.6 (32.7) & \textbf{92.4$\pm$3.5} (45.8) & \textbf{93.0$\pm$3.2} (22.2) & 90.2$\pm$7.2 (22.4) & 88.3$\pm$10.0 (44.7) & \textbf{92.7$\pm$3.5} (47.9) & 91.3$\pm$4.4 (45.1) \\ 
\hline
Ion1 & 29.4$\pm$4.3 (87.6) & 26.5$\pm$3.8 (34.0) & \textbf{54.2$\pm$9.4} (87.2) & 22.7$\pm$8.6 (33.0) & \textbf{56.4$\pm$10.8} (43.0) & 26.7$\pm$5.7 (77.0) & 26.5$\pm$3.8 (34.0) & \textbf{56.1$\pm$8.6} (87.6) \\ 
Ion2 & \textbf{89.0$\pm$1.5} (157.4) & 83.6$\pm$2.7 (121.8) & 84.0$\pm$3.1 (156.4) & 81.7$\pm$5.4 (66.3) & \textbf{84.1$\pm$2.5} (77.4) & 85.0$\pm$1.8 (157.4) & 84.4$\pm$2.9 (157.4) & \textbf{86.0$\pm$3.8} (157.4) \\ 
\hline
Son1 & 53.9$\pm$6.6 (68.0) & \textbf{55.6$\pm$5.0} (64.4) & 53.9$\pm$5.4 (65.3) & \textbf{56.8$\pm$4.2} (33.0) & 51.8$\pm$5.5 (33.0) & \textbf{54.2$\pm$6.1} (67.4) & 53.2$\pm$6.8 (66.6) & 53.7$\pm$7.9 (66.4) \\ 
Son2& 54.5$\pm$9.1 (78.0) & 53.8$\pm$8.2 (64.3) & \textbf{54.7$\pm$8.8} (72.9) & \textbf{56.0$\pm$8.0} (38.0) & 55.6$\pm$5.9 (38.0) & 53.5$\pm$7.9 (67.4) & \textbf{56.8$\pm$4.9} (62.9) & 54.9$\pm$9.2 (72.2) \\ 
\hline
Bank1 & \textbf{95.5$\pm$3.4} (24.0) & 90.3$\pm$4.6 (17.5) & 93.8$\pm$3.3 (23.4) & 87.7$\pm$7.2 (10.0) & \textbf{93.0$\pm$3.8} (22.9) & \textbf{95.2$\pm$2.3} (24.0) & 94.0$\pm$2.6 (24.0) & 93.3$\pm$2.6 (23.6) \\ 
Bank2 & 28.5$\pm$16.2 (67.4) & 15.1$\pm$7.8 (18.9) & \textbf{67.4$\pm$9.1} (67.4) & 14.9$\pm$7.7 (15.5) & \textbf{62.3$\pm$9.0} (49.0) & 35.6$\pm$14.7 (67.4) & 13.7$\pm$8.8 (27.6) & \textbf{68.0$\pm$10.0} (67.4) \\ 
\hline
Happ1 & \textbf{51.6$\pm$8.9} (45.6) & \textbf{51.6$\pm$8.9} (8.0) & 50.4$\pm$8.1 (22.5) & \textbf{52.2$\pm$8.4} (6.6) & 51.8$\pm$8.9 (12.7) & \textbf{53.4$\pm$9.0} (14.0) & 51.6$\pm$8.9 (10.4) & 51.1$\pm$7.5 (22.8) \\ 
Happ2 & \textbf{43.2$\pm$9.0} (44.6) & \textbf{43.2$\pm$9.0} (7.9) & 42.5$\pm$10.0 (11.1) & \textbf{44.2$\pm$9.6} (7.0) & 40.9$\pm$10.8 (10.4) & 42.7$\pm$10.1 (11.2) & 42.6$\pm$9.9 (10.4) & \textbf{44.0$\pm$9.2} (11.3) \\ 
\hline
Aver. & 69.3$\pm$7.2 (58.7) & 67.6$\pm$6.5 (34.7) & \textbf{74.6$\pm$6.0} (49.7) & 68.2$\pm$6.2 (21.9) & \textbf{73.9$\pm$6.4} (28.3) & 70.2$\pm$6.1 (52.3) & 68.6$\pm$5.9 (42.3) & \textbf{74.3$\pm$7.1} (50.6) \\
\hline
\end{tabular}}
\end{center}
\end{table*}

\begin{table*}[htbp]
\caption{Average Gmean values (and feature space dimensions) for \ac{SVDD} with \ac{NPT} using the base kernel function and the seven cases of our proposed generalized reference mapping function}
\label{tab:SVDDnpt}
\begin{center}
\resizebox{1\linewidth}{!}{
\begin{tabular}{|c||c|c|c||c|c||c|c|c|}
\hline
&\multicolumn{3}{|c||}{\textbf{$M=N$}}&\multicolumn{2}{c||}{\textbf{$M=\lfloor N/2 \rfloor$}}&\multicolumn{3}{c|}{\textbf{$M=N+T$}} \\
\cline{2-9} 
\textbf{Data} & \textbf{Base} & \textbf{Case 1} & \textbf{Case 2} & \textbf{Case 3} & \textbf{Case 4} & \textbf{Case 5} & \textbf{Case 6} & \textbf{Case 7}  \\
\hline
Iris1 & 89.8$\pm$4.1 (14.5) & 89.8$\pm$4.1 (14.5) & \textbf{91.3$\pm$3.7} (12.8) & \textbf{91.8$\pm$5.5} (9.1) & 88.3$\pm$8.7 (10.0) & 90.8$\pm$3.8 (35.8) & 91.1$\pm$3.8 (17.6) & \textbf{91.9$\pm$3.6} (15.2) \\ 
Iris2 & 90.7$\pm$4.9 (5.0) & 90.7$\pm$4.9 (5.0) & \textbf{90.9$\pm$4.9} (6.2) & \textbf{91.3$\pm$4.8} (5.7) & 90.4$\pm$4.7 (4.6) & \textbf{91.4$\pm$5.6} (12.2) & 90.9$\pm$4.6 (9.5) & 90.9$\pm$4.6 (10.2) \\ 
Iris3 & 89.4$\pm$3.1 (18.9) & 89.4$\pm$3.1 (18.9) & \textbf{89.9$\pm$3.5} (25.5) & \textbf{90.4$\pm$2.8} (12.8) & 90.0$\pm$3.4 (14.4) & 88.4$\pm$5.3 (37.8) & 89.8$\pm$3.6 (33.1) & \textbf{90.2$\pm$3.1} (28.1) \\ 
\hline
Seed1 & 83.1$\pm$3.9 (15.0) & 83.1$\pm$3.9 (15.0) & \textbf{84.3$\pm$4.1} (31.5) & 83.8$\pm$3.4 (11.6) & \textbf{83.9$\pm$3.6} (17.4) & \textbf{86.4$\pm$2.6} (61.7) & 83.1$\pm$3.3 (41.6) & 83.6$\pm$4.8 (33.8) \\ 
Seed2 & \textbf{90.9$\pm$4.8} (21.0) & \textbf{90.9$\pm$4.8} (21.0) & 90.8$\pm$4.9 (31.0) & 89.7$\pm$4.3 (13.9) & \textbf{90.8$\pm$4.7} (16.3) & \textbf{93.1$\pm$3.6} (33.1) & 91.1$\pm$5.1 (38.4) & 91.0$\pm$4.7 (66.6) \\ 
Seed3 & \textbf{93.6$\pm$3.5} (11.7) & \textbf{93.6$\pm$3.5} (11.7) & 93.5$\pm$3.4 (10.7) & \textbf{93.8$\pm$3.3} (9.7) & 93.6$\pm$3.5 (10.1) & \textbf{93.7$\pm$3.4} (32.9) & 93.6$\pm$3.4 (15.0) & 93.6$\pm$3.4 (14.1) \\ 
\hline
Ion1 & 33.0$\pm$18.5 (34.0) & 33.0$\pm$18.5 (34.0) & \textbf{41.2$\pm$13.2} (87.2) & 32.2$\pm$18.2 (33.0) & \textbf{44.4$\pm$19.1} (43.0) & 34.0$\pm$10.7 (85.4) & 33.0$\pm$18.5 (34.0) & \textbf{46.8$\pm$8.5} (175.4) \\ 
Ion2 & \textbf{86.4$\pm$1.8} (105.4) & \textbf{86.4$\pm$1.8} (105.4) & 86.1$\pm$2.1 (156.4) & 85.1$\pm$2.6 (53.9) & \textbf{86.3$\pm$2.1} (77.4) & 85.8$\pm$2.2 (226.8) & \textbf{86.5$\pm$1.6} (178.8) & 86.2$\pm$2.0 (245.0) \\ 
\hline
Son1 & 50.5$\pm$8.6 (65.7) & 50.5$\pm$8.6 (65.7) & \textbf{50.9$\pm$7.8} (67.0) & \textbf{51.2$\pm$7.3} (33.0) & 50.3$\pm$7.9 (33.0) & \textbf{51.5$\pm$6.4} (134.2) & 51.4$\pm$6.1 (134.2) & 50.0$\pm$7.8 (135.0) \\ 
Son2 & 56.5$\pm$9.2 (76.6) & 56.5$\pm$9.2 (76.6) & \textbf{57.1$\pm$8.7} (77.0) & \textbf{55.4$\pm$8.5} (38.0) & 54.3$\pm$7.6 (38.0) & \textbf{57.2$\pm$10.1} (135.8) & 55.9$\pm$9.5 (140.1) & 55.0$\pm$9.2 (145.0) \\
\hline
Bank1 & 93.4$\pm$3.3 (17.6) & 93.5$\pm$3.1 (17.6) & \textbf{94.6$\pm$2.7} (56.4) & 94.3$\pm$2.9 (13.2) & \textbf{95.4$\pm$2.8} (34.6) & 95.3$\pm$2.5 (72.3) & 94.6$\pm$3.0 (53.8) & \textbf{96.1$\pm$2.9} (76.4) \\ 
Bank2 & 32.9$\pm$27.6 (9.6) & 32.9$\pm$27.6 (9.6) & \textbf{60.6$\pm$15.6} (99.4) & 31.0$\pm$27.1 (6.0) & \textbf{57.2$\pm$13.9} (49.0) & 38.1$\pm$5.0 (90.4) & 32.9$\pm$27.6 (15.4) & \textbf{68.4$\pm$12.7} (174.0) \\ 
\hline
Happ1 & 39.5$\pm$15.8 (20.9) & 39.5$\pm$15.8 (20.9) & \textbf{42.8$\pm$9.4} (47.0) & 39.2$\pm$17.0 (16.0) & \textbf{41.8$\pm$16.5} (20.2) & \textbf{45.1$\pm$10.4} (62.0) & 41.7$\pm$9.2 (66.4) & 44.1$\pm$10.0 (67.4) \\ 
Happ2 & \textbf{47.9$\pm$13.3} (9.4) & \textbf{47.9$\pm$13.3} (9.4) & 44.7$\pm$11.8 (20.5) & 40.4$\pm$12.4 (11.4) & \textbf{41.2$\pm$11.4} (10.2) & 43.0$\pm$11.3 (34.7) & \textbf{44.8$\pm$11.6} (33.3) & 40.9$\pm$12.4 (25.0) \\ 
\hline
Aver. & 69.8$\pm$8.7 (30.4) & 69.8$\pm$8.7 (30.4) & \textbf{72.8$\pm$6.8} (52.0) & 69.3$\pm$8.6 (19.1) & \textbf{72.0$\pm$7.8} (27.0) & 71.0$\pm$5.9 (75.4) & 70.0$\pm$7.9 (57.9) & \textbf{73.5$\pm$6.4} (86.5) \\
\hline
\end{tabular}}
\end{center}
\end{table*}

\begin{table*}[htbp]
\caption{Average Gmean values (and kernel matrix ranks) for \ac{OCSVM} with the base kernel function and the seven cases of our proposed generalized reference kernel function}
\label{tab:OCSVMkernel}
\begin{center}
\resizebox{1\linewidth}{!}{
\begin{tabular}{|c||c|c|c||c|c||c|c|c|}
\hline
&\multicolumn{3}{|c||}{\textbf{$M=N$}}&\multicolumn{2}{c||}{\textbf{$M=\lfloor N/2 \rfloor$}}&\multicolumn{3}{c|}{\textbf{$M=N+T$}} \\
\cline{2-9} 
\textbf{Data} & \textbf{Base} & \textbf{Case 1} & \textbf{Case 2} & \textbf{Case 3} & \textbf{Case 4} & \textbf{Case 5} & \textbf{Case 6} & \textbf{Case 7}  \\
\hline
Iris1 & \textbf{88.6$\pm$4.4} (34.3) & 62.5$\pm$22.6 (13.8) & 77.3$\pm$7.7 (32.7) & 58.7$\pm$26.0 (9.8) & \textbf{72.8$\pm$23.4} (16.0) & \textbf{95.1$\pm$5.0} (18.5) & 64.3$\pm$24.4 (23.8) & 70.9$\pm$7.4 (34.7) \\ 
Iris2 & \textbf{91.1$\pm$4.2} (30.6) & 30.0$\pm$22.5 (25.9) & 76.1$\pm$9.0 (34.0) & 47.9$\pm$17.1 (13.8) & \textbf{79.2$\pm$7.6} (16.0) & \textbf{89.4$\pm$5.4} (35.0) & 69.0$\pm$6.9 (35.0) & 72.1$\pm$9.1 (35.0) \\ 
Iris3& \textbf{90.2$\pm$3.2} (34.8) & 60.3$\pm$15.5 (13.1) & 76.1$\pm$11.7 (34.0) & 43.8$\pm$21.1 (9.0) & \textbf{74.4$\pm$10.6} (16.0) & \textbf{93.7$\pm$2.0} (19.2) & 69.2$\pm$16.3 (30.6) & 65.3$\pm$22.7 (26.4) \\ 
\hline  
Seed1& \textbf{86.5$\pm$2.1} (45.8) & 49.4$\pm$23.9 (24.2) & 74.2$\pm$7.3 (48.0) & 43.1$\pm$18.4 (12.8) & \textbf{75.6$\pm$6.7} (23.0) & \textbf{86.7$\pm$5.3} (42.2) & 70.8$\pm$6.8 (49.0) & 74.8$\pm$6.1 (49.0) \\ 
Seed2 & \textbf{89.6$\pm$4.4} (48.2) & 59.1$\pm$27.8 (36.8) & 82.0$\pm$7.3 (48.0) & 56.1$\pm$22.0 (17.9) & \textbf{81.1$\pm$7.5} (23.0) & \textbf{88.7$\pm$5.4} (23.2) & 63.5$\pm$16.7 (42.9) & 81.8$\pm$6.9 (49.0) \\ 
Seed3 & \textbf{92.3$\pm$4.4} (48.9) & 53.9$\pm$28.7 (31.2) & 78.7$\pm$7.0 (48.0) & 36.7$\pm$27.8 (15.4) & \textbf{83.2$\pm$7.5} (23.0) & \textbf{89.5$\pm$7.7} (39.0) & 72.9$\pm$9.2 (49.0) & 82.3$\pm$6.3 (49.0) \\ 
\hline
Ion1& 36.7$\pm$2.6 (87.6) & \textbf{51.0$\pm$9.8} (84.1) & 48.5$\pm$14.0 (85.0) & \textbf{49.2$\pm$10.0} (42.5) & 47.3$\pm$9.9 (42.2) & \textbf{70.6$\pm$4.0} (87.6) & 51.1$\pm$11.4 (79.1) & 52.9$\pm$10.1 (87.6) \\ 
Ion2 & \textbf{88.9$\pm$2.3} (157.4) & 47.0$\pm$24.9 (34.0) & 76.5$\pm$5.2 (156.4) & 50.9$\pm$16.6 (48.5) & \textbf{67.4$\pm$14.8} (77.4) & \textbf{88.2$\pm$3.0} (157.4) & 79.3$\pm$2.7 (157.4) & 79.7$\pm$4.3 (157.4) \\ 
\hline
Son1 & \textbf{53.2$\pm$5.7} (68.0) & 47.2$\pm$5.7 (67.0) & 50.2$\pm$6.8 (67.0) & 46.7$\pm$11.4 (33.0) & \textbf{50.1$\pm$6.3} (33.0) & \textbf{63.9$\pm$4.3} (68.0) & 46.9$\pm$5.8 (67.2) & 53.7$\pm$5.6 (68.0) \\ 
Son2 & \textbf{55.1$\pm$8.8} (78.0) & 50.8$\pm$6.7 (77.0) & 51.4$\pm$9.0 (77.0) & 48.5$\pm$9.2 (38.0) & \textbf{48.9$\pm$9.3} (38.0) & \textbf{67.9$\pm$4.8} (76.6) & 51.7$\pm$8.6 (78.0) & 51.7$\pm$6.6 (78.0) \\ 
\hline
Bank1 & \textbf{93.0$\pm$3.4} (23.4) & 62.1$\pm$3.6 (10.2) & 69.9$\pm$9.7 (24.0) & 58.6$\pm$14.2 (18.1) & \textbf{69.5$\pm$8.5} (24.0) & \textbf{86.0$\pm$4.4} (24.0) & 69.3$\pm$10.0 (24.0) & 70.0$\pm$8.5 (24.0) \\ 
Bank2 & 20.7$\pm$12.0 (67.4) & 57.4$\pm$6.8 (57.9) & \textbf{80.9$\pm$9.0} (67.4) & 57.3$\pm$14.4 (37.0) & \textbf{82.3$\pm$8.2} (49.0) & 78.3$\pm$4.3 (65.8) & 59.4$\pm$21.3 (67.4) & \textbf{82.2$\pm$4.8} (67.4) \\ 
\hline
Happ1 & \textbf{54.6$\pm$7.1} (47.0) & 44.9$\pm$10.9 (36.5) & 45.2$\pm$9.2 (36.0) & \textbf{48.6$\pm$8.4} (23.3) & 48.5$\pm$8.3 (20.6) & \textbf{52.7$\pm$8.6} (45.9) & 47.6$\pm$6.9 (41.2) & 47.5$\pm$7.8 (37.1) \\ 
Happ2 & 44.3$\pm$8.0 (44.6) & \textbf{47.9$\pm$7.8} (38.2) & 42.9$\pm$9.2 (34.4) & 42.3$\pm$6.0 (17.9) & \textbf{43.4$\pm$8.7} (17.4) & 45.5$\pm$9.4 (32.7) & \textbf{48.0$\pm$9.6} (26.0) & 42.4$\pm$12.9 (27.0) \\ 
\hline
Aver. & \textbf{70.3$\pm$5.2} (58.3) & 51.7$\pm$15.5 (39.3) & 66.4$\pm$8.7 (56.6) & 49.2$\pm$15.9 (24.1) & \textbf{66.0$\pm$9.8} (29.9) & \textbf{78.3$\pm$5.3} (52.5) & 61.6$\pm$11.2 (55.0) & 66.2$\pm$8.5 (56.4) \\ 
\hline
\end{tabular}}
\end{center}
\end{table*}

\begin{table*}[htbp]
\caption{Average Gmean values (and feature space dimensions) for \ac{OCSVM} with \ac{NPT} using the base kernel function and the seven cases of our proposed generalized reference mapping function}
\label{tab:OCSVMnpt}
\begin{center}
\resizebox{1\linewidth}{!}{
\begin{tabular}{|c||c|c|c||c|c||c|c|c|}
\hline
&\multicolumn{3}{|c||}{\textbf{$M=N$}}&\multicolumn{2}{c||}{\textbf{$M=\lfloor N/2 \rfloor$}}&\multicolumn{3}{c|}{\textbf{$M=N+T$}} \\
\cline{2-9} 
\textbf{Data} & \textbf{Base} & \textbf{Case 1} & \textbf{Case 2} & \textbf{Case 3} & \textbf{Case 4} & \textbf{Case 5} & \textbf{Case 6} & \textbf{Case 7}  \\
\hline
Iris1 & 62.5$\pm$22.6 (13.8) & 62.5$\pm$22.6 (13.8) & \textbf{77.3$\pm$7.7} (32.7) & 58.7$\pm$26.0 (9.8) & \textbf{72.8$\pm$23.4} (16.0) & \textbf{95.1$\pm$5.0} (19.3) & 64.3$\pm$24.4 (32.6) & 70.9$\pm$7.4 (46.8) \\ 
Iris2 & 30.0$\pm$22.5 (25.9) & 30.0$\pm$22.5 (25.9) & \textbf{76.1$\pm$9.0} (34.0) & 47.9$\pm$17.1 (13.8) & \textbf{79.2$\pm$7.6} (16.0) & \textbf{89.4$\pm$5.4} (50.6) & 69.0$\pm$6.9 (67.2) & 72.1$\pm$9.1 (68.5) \\ 
Iris3 & 60.3$\pm$15.5 (13.1) & 60.3$\pm$15.5 (13.1) & \textbf{76.1$\pm$11.7} (34.0) & 43.8$\pm$21.1 (9.0) & \textbf{74.4$\pm$10.6} (16.0) & \textbf{93.7$\pm$2.0} (19.2) & 69.2$\pm$16.3 (54.1) & 65.3$\pm$22.7 (49.1) \\ 
\hline
Seed1 & 49.4$\pm$23.9 (24.2) & 49.4$\pm$23.9 (24.2) & \textbf{74.2$\pm$7.3} (48.0) & 43.1$\pm$18.4 (12.8) & \textbf{75.6$\pm$6.7} (23.0) & \textbf{86.7$\pm$5.3} (52.1) & 70.8$\pm$6.8 (97.0) & 74.8$\pm$6.1 (97.0) \\ 
Seed2 & 59.1$\pm$27.8 (36.8) & 59.1$\pm$27.8 (36.8) & \textbf{82.0$\pm$7.3}(48.0) & 56.1$\pm$22.0 (17.9) & \textbf{81.1$\pm$7.5} (23.0) & \textbf{88.7$\pm$5.4} (23.2) & 63.5$\pm$16.7 (83.2) & 81.8$\pm$6.9 (97.0) \\ 
Seed3& 53.9$\pm$28.7 (31.2) & 53.9$\pm$28.7 (31.2) & \textbf{78.7$\pm$7.0} (48.0) & 36.7$\pm$27.8 (15.4) & \textbf{83.2$\pm$7.5} (23.0) & \textbf{89.5$\pm$7.7} (49.2) & 72.9$\pm$9.2 (97.0) & 82.3$\pm$6.3 (97.0) \\
\hline
Ion1 & 51.0$\pm$9.8 (84.1) & \textbf{51.2$\pm$9.8} (84.1) & 48.5$\pm$14.0 (85.0) & \textbf{49.2$\pm$10.0} (42.5) & 47.3$\pm$9.9 (42.2) & \textbf{70.6$\pm$4.0} (151.0) & 51.1$\pm$11.4 (140.2) & 52.9
$\pm$10.1 (175.4) \\ 
Ion2 & 47.0$\pm$24.9 (34.0) & 47.0$\pm$24.9 (34.0) & \textbf{76.5$\pm$5.2} (156.4) & 50.9$\pm$16.6 (48.5) & \textbf{67.4$\pm$14.8} (77.4) & \textbf{88.2$\pm$3.0} (238.6) & 79.3$\pm$2.7 (244.8) & 79.7$\pm$4.3 (245.0) \\
\hline
Son1 & 47.2$\pm$5.7 (67.0) & 47.2$\pm$5.7 (67.0) & \textbf{50.2$\pm$6.8} (67.0) & 46.7$\pm$11.4 (33.0) & \textbf{50.1$\pm$6.3} (33.0) & \textbf{63.9$\pm$4.3} (132.3) & 46.9$\pm$5.8 (125.0) & 53.7$\pm$5.6 (135.0) \\ 
Son2 & 50.8$\pm$6.7 (77.0) & 50.8$\pm$6.7 (77.0) & \textbf{51.4$\pm$9.0} (77.0) & 48.5$\pm$9.2 (38.0) & \textbf{48.9$\pm$9.3} (38.0) & \textbf{67.9$\pm$4.8} (135.9) & 51.7$\pm$8.6 (140.2) & 51.7$\pm$6.6 (145.0) \\
\hline
Bank1 & 62.1$\pm$4.2 (10.8) & 61.6$\pm$3.7 (10.2) & \textbf{70.1$\pm$9.5} (73.4) & 59.6$\pm$14.3 (18.1) & \textbf{68.4$\pm$9.0} (36.0) & \textbf{86.0$\pm$4.4} (75.2) & 69.3$\pm$10.0 (97.4) & 69.8$\pm$7.5 (147.8) \\ 
Bank2& 58.9$\pm$8.3 (60.7) & 58.0$\pm$5.7 (57.9) & \textbf{80.9$\pm$9.0} (99.4) & 57.3$\pm$14.5 (37.0) & \textbf{82.4$\pm$8.1} (49.0) & 78.5$\pm$6.7 (82.7) & 59.4$\pm$21.3 (140.4) & \textbf{82.2$\pm$4.8} (174.0) \\ 
\hline
Happ1 & 45.3$\pm$11.0 (36.5) & 44.9$\pm$10.9 (36.5) & \textbf{45.4$\pm$9.1} (38.8) & \textbf{48.6$\pm$8.4} (23.3) & 48.5$\pm$8.3 (20.6) & \textbf{52.7$\pm$8.6} (69.5) & 47.6$\pm$6.9 (66.9) & 47.5$\pm$7.8 (65.5) \\ 
Happ2 & \textbf{48.4$\pm$7.3} (38.9) & 47.9$\pm$7.8 (38.2) & \textbf{42.9$\pm$9.2} (34.8) & 42.3$\pm$6.0 (17.9) & 43.4$\pm$8.7 (17.4) & 45.5$\pm$9.4 (40.0) & \textbf{48.0$\pm$9.6} (35.5) & 42.4$\pm$12.9 (36.8) \\ 
\hline
Aver. & 51.9$\pm$15.6 (39.6) & 51.7$\pm$15.4 (39.3) & \textbf{66.4$\pm$8.7} (62.6) & 49.2$\pm$15.9 (24.1) & \textbf{65.9$\pm$9.8} (30.8) & \textbf{78.3$\pm$5.4} (81.4) & 61.6$\pm$11.2 (101.5) & 66.2$\pm$8.4 (112.8) \\ 
\hline
\end{tabular}}
\end{center}
\end{table*}

The experimental results in terms of Gmean are given in Table~\ref{tab:SVDDkernel}-\ref{tab:OCSVMnpt}. Tables~\ref{tab:SVDDkernel} shows \ac{SVDD} results using the base kernel matrix along with the seven cases of our proposed generalized kernel matrix \eqref{eq:refkernelmatrix} via kernel implementation and Table \ref{tab:SVDDnpt} shows the results using \ac{NPT} along with the seven cases of our proposed generalized kernel mapping  \eqref{eq:refmapping} via linear implementation. Tables~\ref{tab:OCSVMkernel} and Tables~\ref{tab:OCSVMnpt} have the corresponding results for \ac{OCSVM}. The tables also give the average kernel matrix ranks or feature space dimensions for the mappings. In each table, the best performance for each number of reference vectors is bolded.

First, we see that the results for the base \ac{NPT} and for the case 1 of the proposed mapping are indeed almost identical. For the kernel setting, the difference is larger in particular for \ac{OCSVM}. Our additional experiments verify that the differences indeed follow mainly from the centering operation and, in most cases, \ac{OCSVM} suffers from the centering, while for \ac{SVDD} centering slightly improves the results. Also in general, we see that the conclusions for \ac{SVDD} and \ac{OCSVM} are quite different, which shows that future analysis should focus on the kernel properties with respect to a specific algorithm.
 
For \ac{SVDD}, the random cases 2, 4, and 7 quite consistently improve the results compared to the base method or to using training vectors as reference vectors. In particular, there are few cases (Ion1, Bank2), where the base methods clearly fail, but the proposed kernel and mapping with random reference vectors leads to clearly improved results. The random reference vectors indeed seem to provide implicit regularization and, thus, a more robust performance. The results are on average slightly better when there are more reference vectors, but the differences are quite insignificant. Thus, using the proposed generalized reference kernel as an implicit data augmentation ($M>N$) may slightly help, but it is more beneficial to replace the training samples with random data than to augment the training samples with random data (case~6).

Looking at the kernel matrix ranks, we see that the proposed implementation with the case 2 often leads to lower ranks but better results than the original method. On the hand, in the tasks where the random reference vectors lead to significant improvements do not have lower rank. This gives some indication that the random reference vector variant can adapt the rank to the task at hand. For the base \ac{NPT}, ranks are lower than for the base kernel, whereas for the proposed approach the ranks are more similar between the kernel and mapping, while for the mapping the variance in ranks is higher.   

For \ac{OCSVM}, as mentioned already, centering clearly harms the performance. Thus, the base kernel performs much better than the base \ac{NPT}, while most of the proposed generalized reference kernel variants also achieve a lower performance than the corresponding variant for \ac{SVDD}. A very interesting exception for both generalized reference kernel and mapping with \ac{OCSVM} is the case 5 that uses also negative samples as reference vectors. The case 5 results are consistently outperforming all the other cases and also all the \ac{SVDD} results.  While it should be remembered that comparing this case directly with the other cases is unfair, because the basic assumption of not having negative examples is violated, the results still open interesting future work opportunities either as using the case 5 as a way to exploit some negative samples within the regular one-class implementation or as an indication that for \ac{OCSVM} it may be more beneficial to pick reference vectors outside the training data distribution. Furthermore, it is interesting that only \ac{OCSVM} with generalized reference kernel and mapping can exploit the negative data efficiently, while the same is not true for \ac{SVDD} with the proposed approaches. Further analyzing this difference may lead to better understanding how to select most suitable kernels for different algorithms.

\section{Conclusions}
\label{sec:conclusion}

In this paper, we proposed a new kernel formulation based on a base kernel and reference vectors that can be selected in different ways. Different cases (i.e., different ways to select the reference vectors) of our \emph{generalized reference mapping and kernel} have links to different prior works including approximate kernels and \ac{NPT}. Our formulation also provides a way to extend random sampling and random projection-based approximate kernel methods into an \ac{NPT}-like setting. In this paper, we considered the new formulation in the context of small-scale one-class classification and with \ac{RBF} kernels, but in the future it can be used also in different tasks and with different kernels.

Our experimental results show that the new formulation can help to implicitly regularize and adjust the rank of the kernel matrices. It also allows to incorporate additional information into the kernel itself, leading to improved one-class classification accuracy. For \ac{SVDD}, random reference vectors led to best classification results and more robust performance indicating that this approach provides implicit regularization. For \ac{OCSVM}, using negative samples as references vectors led to significant classification performance improvements. While this approach violates the assumption that one-class classification cannot use negative examples for training, it can be considered in the future in the class-specific classification context or in one-class classification with few negative examples. It may also indicate that for \ac{OCSVM} it would be more beneficial to select random reference vector outside the training distribution.

As the conclusions for \ac{SVDD} and \ac{OCSVM} are quite different, this suggests that the new kernel formulation should be further studied with respect to a specific method. On the other hand, further analyzing the causes for the observed differences may help to understand how to optimize the reference vectors for the method and task at hand.

\appendix
\label{appendix}
\setcounter{equation}{0}
\numberwithin{equation}{section}
In NPT, the training data needs to be centered in $\mathcal{F}$, i.e. $\sum_{i=1}^\NoItems \phi(\input_i) = 0$. If this assumption is not met, $\mathcal{P}$ is a manifold in $\mathcal{F}$ and not a subspace of $\mathcal{F}$, since $\mathbf{0}$ does not necessarily belong is to $\mathcal{P}$ \cite{kwak2013NPT}. In our formulation, $\mathcal{F}$ is spanned by function $\phi(\cdot)$, meaning that the centering operation is not mandatory. Nevertheless, in practice we apply \ac{KPCA} for the reference data with the base kernel $\bkernelfunction(\cdot,\cdot)$ and, therefore, we also opt to center our reference data in the base kernel space $\tilde{\mathcal{F}}$ by centering the corresponding kernel function $\bkernelmatrix_{\Refs \Refs}$. To align our training and testing data, we also center $\bkernelmatrix_{\Refs \Input}$ and $\bkernelmatrix_{\Refs \hat{\Input}}$ with respect to the center of the reference data $\Refs$.

Let us denote by $\tilde{\boldsymbol{\mu}}$ the mean of the $\noRefs$ uncentered reference vectors $\tilde{\Phi}(\Refs) = [\tilde{\phi}(\refs_1), \dots, \tilde{\phi}(\refs_\noRefs)] \in \tilde{\mathcal{F}}$:
\begin{equation}
\tilde{\boldsymbol{\mu}} = \frac{1}{\noRefs} \sum_{i=1}^\noRefs \tilde{\phi}(\refs_i) = \frac{1}{\noRefs} \tilde{\Phi}(\Refs) \one_{\noRefs},
\end{equation}
where $\one_{\noRefs} \in \rspace^{\noRefs} $ is a vector of ones. The centered reference vectors in $\tilde{\mathcal{F}}$ are given by
\begin{equation}
\tilde{\Psi}(\Refs) = \tilde{\Phi}(\Refs) - \tilde{\boldsymbol{\mu}} \mathbf{1}_{\noRefs}^T = \tilde{\Phi}(\Refs)(\Identity - \frac{1}{\noRefs}\one_{\noRefs}\one_{\noRefs}^T) = \tilde{\Phi}\mathbf{C}_{\noRefs},
\end{equation}
where we denote by $\mathbf{C}_{\noRefs} = \Identity - \frac{1}{\noRefs}\one_{\noRefs}\one_{\noRefs}^T$ the centering matrix for the reference vectors. The training vectors $\Input$ (and test vectors $\hat{\Input}$ accordingly) can be centered with respect to $\tilde{\boldsymbol{\mu}}$ in $\tilde{\mathcal{F}}$ as follows:
\begin{eqnarray}
\tilde{\Psi}(\Input) &=& \tilde{\Phi}(\Input) - \tilde{\boldsymbol{\mu}} \mathbf{1}_{\NoItems}^T = \tilde{\Phi}(\Input) - \frac{1}{\noRefs} \tilde{\Phi}(\Refs) \mathbf{1}_{\noRefs} \mathbf{1}_{\NoItems}^T \nonumber \\
&=& \tilde{\Phi}(\Input) - \tilde{\Phi}(\Refs)\mathbf{C}_{\NoItems}, 
\end{eqnarray}
where $\mathbf{C}_{\NoItems} = \frac{1}{\noRefs} \mathbf{1}_{\noRefs} \mathbf{1}_{\NoItems}^T$.
By using $\tilde{\Psi}(\Refs)$ and $\tilde{\Psi}(\Input)$ (or $\tilde{\Psi}(\hat{\Input})$), the centered kernel matrices $\tilde{\mathcal{K}}_{\Refs \Refs}$ and $\tilde{\mathcal{K}}_{\Refs \Input}$ (or $\tilde{\mathcal{K}}_{\Refs \hat{\Input}}$) can be calculated as follows:
\begin{equation}
\label{eq:center_ref}
\tilde{\mathcal{K}}_{\Refs \Refs} = \tilde{\Psi}(\Refs)^T \tilde{\Psi}(\Refs) = \mathbf{C}_{\noRefs} \tilde{\kernelmatrix}_{\Refs \Refs} \mathbf{C}_{\noRefs}
\end{equation}
and
\begin{equation}
\label{eq:center_other}
\tilde{\mathcal{K}}_{\Refs \Input} = \tilde{\Psi}(\Refs)^T \tilde{\Psi}(\Input) = \mathbf{C}_{\noRefs} \left( \tilde{\kernelmatrix}_{\Refs \Refs} - \tilde{\kernelmatrix}_{\Refs \Input} \mathbf{C}_{\NoItems} \right),
\end{equation}
Note that the centering formulas for \ac{NPT} can be obtained by setting $\Refs = \Input$ and by using the kernel at hand as the base kernel in the formulas.
\bibliographystyle{IEEEtran}
\bibliography{references}

\end{document}